%% file: main.tex
\documentclass[10pt,twocolumn]{article}

\usepackage[letterpaper,margin=0.72in]{geometry}
\usepackage[T1]{fontenc}
\usepackage[utf8]{inputenc}
\usepackage{times}
\usepackage{microtype}
\usepackage{graphicx}
\usepackage{booktabs}
\usepackage{amsmath}
\usepackage{amssymb}
\usepackage{caption}
\usepackage[numbers,sort&compress]{natbib}
\usepackage[colorlinks=true,citecolor=blue,linkcolor=blue,urlcolor=blue]{hyperref}
\usepackage{balance}
\usepackage{stfloats}
\usepackage{xcolor}

\graphicspath{{figures/}}
\input{generated/numbers}

\makeatletter\@secpenalty=0\makeatother 
\title{\vspace{-1.0em}\textbf{The Shape of Wisdom:\\How Language Models Converge on an Answer}}
\author{Shailesh Rana\\Independent Researcher}
\date{}

\begin{document}

\twocolumn[
\begin{@twocolumnfalse}
\maketitle
\vspace{-1.4em}
\begin{center}
\begin{minipage}{0.82\textwidth}
\begin{abstract}
\input{sections/abstract}
\end{abstract}
\end{minipage}
\end{center}
\vspace{0.7em}
\end{@twocolumnfalse}
]

\input{sections/intro}
\input{sections/setup}
\input{sections/results}
\input{sections/related}
\input{sections/limits}
\input{sections/conclusion}

\paragraph{Code and data.} Code, the dataset and all derived results: \url{https://github.com/gut-puncture/The-Shape-of-Wisdom}.

\paragraph{Acknowledgements.} \input{sections/ack}

\small
\bibliographystyle{plainnat}
\bibliography{references}

\small
\appendix
\input{sections/appendix}

\end{document}

%% file: generated/numbers.tex
\newcommand{\AccGemma}{67\%}
\newcommand{\AccLlama}{63\%}
\newcommand{\AccLlamaSmall}{48\%}
\newcommand{\AccQwen}{58\%}

\newcommand{\AucConfMax}{0.69}
\newcommand{\AucConfMeanGemma}{0.49}
\newcommand{\AucConfMeanLlama}{0.66}
\newcommand{\AucConfMeanLlamaSmall}{0.66}
\newcommand{\AucConfMeanMax}{0.66}
\newcommand{\AucConfMeanMin}{0.49}
\newcommand{\AucConfMeanQwen}{0.53}
\newcommand{\AucConfMin}{0.50}

\newcommand{\AucGeoMax}{0.83}
\newcommand{\AucGeoMin}{0.73}

\newcommand{\AucOutRankGemma}{0.80}
\newcommand{\AucOutRankLlama}{0.82}
\newcommand{\AucOutRankLlamaSmall}{0.80}
\newcommand{\AucOutRankMax}{0.82}
\newcommand{\AucOutRankMin}{0.77}
\newcommand{\AucOutRankQwen}{0.77}
\newcommand{\AucRankGemma}{0.75}
\newcommand{\AucRankLlama}{0.77}
\newcommand{\AucRankLlamaSmall}{0.82}
\newcommand{\AucRankMax}{0.82}
\newcommand{\AucRankMin}{0.75}
\newcommand{\AucRankQwen}{0.75}
\newcommand{\CountryAucConfMax}{0.74}
\newcommand{\CountryAucConfMin}{0.36}
\newcommand{\CountryAucRankMax}{0.84}
\newcommand{\CountryAucRankMin}{0.74}
\newcommand{\CrossModelMax}{88\%}
\newcommand{\CrossModelMin}{66\%}
\newcommand{\CrossModelPairs}{12}
\newcommand{\EarlyMax}{55\%}
\newcommand{\EarlyMin}{20\%}

\newcommand{\ErrPosGemma}{52\%}
\newcommand{\ErrPosLlama}{60\%}
\newcommand{\ErrPosLlamaSmall}{6\%}

\newcommand{\ErrPosQwen}{54\%}
\newcommand{\GivenLateGemma}{92\%}
\newcommand{\GivenLateLlama}{94\%}
\newcommand{\GivenLateLlamaSmall}{86\%}
\newcommand{\GivenLateMax}{94\%}
\newcommand{\GivenLateMin}{86\%}
\newcommand{\GivenLateQwen}{90\%}
\newcommand{\GoldLateGemma}{73\%}
\newcommand{\GoldLateLlama}{75\%}
\newcommand{\GoldLateLlamaSmall}{61\%}
\newcommand{\GoldLateMax}{75\%}
\newcommand{\GoldLateMin}{61\%}
\newcommand{\GoldLateQwen}{65\%}

\newcommand{\JumpFromGemma}{35\%}
\newcommand{\JumpFromLlama}{32\%}
\newcommand{\JumpFromLlamaSmall}{35\%}
\newcommand{\JumpFromMax}{35\%}
\newcommand{\JumpFromMin}{28\%}
\newcommand{\JumpFromQwen}{28\%}

\newcommand{\JumpLayerBeforeLlama}{14}

\newcommand{\JumpToGemma}{83\%}
\newcommand{\JumpToLlama}{71\%}
\newcommand{\JumpToLlamaSmall}{74\%}
\newcommand{\JumpToQwen}{67\%}
\newcommand{\LayersGemma}{42}
\newcommand{\LayersLlama}{32}
\newcommand{\LayersLlamaSmall}{16}
\newcommand{\LayersQwen}{28}

\newcommand{\MatchPostRightMax}{16\%}
\newcommand{\MatchPostRightMin}{4\%}
\newcommand{\MatchPostSizeMax}{8\%}
\newcommand{\MatchPostSizeMin}{6\%}
\newcommand{\MatchPreRightMax}{1\%}
\newcommand{\MatchPreRightMin}{0\%}
\newcommand{\MatchPreSizeMax}{8\%}
\newcommand{\MatchPreSizeMin}{5\%}

\newcommand{\NFacts}{2,256}

\newcommand{\NPromptsAnalysed}{15,603}

\newcommand{\NRewrites}{4,323}

\newcommand{\NamesJumpMax}{52}
\newcommand{\NamesJumpMin}{41}

\newcommand{\NamesLateLlama}{94\%}

\newcommand{\NamesLateMax}{94\%}
\newcommand{\NamesLateMin}{89\%}

\newcommand{\NoWordsGemma}{91\%}
\newcommand{\NoWordsLlama}{94\%}
\newcommand{\NoWordsLlamaSmall}{85\%}
\newcommand{\NoWordsQwen}{90\%}
\newcommand{\OnsetGemma}{62\%}
\newcommand{\OnsetLayerGemma}{26}
\newcommand{\OnsetLayerLlama}{15}
\newcommand{\OnsetLayerLlamaSmall}{10}
\newcommand{\OnsetLayerQwen}{20}
\newcommand{\OnsetLlama}{47\%}
\newcommand{\OnsetLlamaSmall}{62\%}
\newcommand{\OnsetMax}{71\%}
\newcommand{\OnsetMin}{47\%}
\newcommand{\OnsetQwen}{71\%}
\newcommand{\PeakGemma}{92\%}
\newcommand{\PeakLlama}{94\%}
\newcommand{\PeakLlamaSmall}{86\%}
\newcommand{\PeakQwen}{92\%}
\newcommand{\PushAllPreMax}{1\%}
\newcommand{\PushAllPreMin}{0\%}

\newcommand{\PushGoldGemma}{97\%}
\newcommand{\PushGoldLlama}{97\%}
\newcommand{\PushGoldLlamaSmall}{83\%}
\newcommand{\PushGoldMax}{97\%}
\newcommand{\PushGoldMin}{83\%}
\newcommand{\PushGoldQwen}{94\%}

\newcommand{\PushNMax}{233}
\newcommand{\PushNMin}{121}

\newcommand{\PushOtherMax}{95\%}
\newcommand{\PushOtherMin}{82\%}

\newcommand{\PushRandomMax}{4\%}
\newcommand{\PushRandomMin}{2\%}

\newcommand{\PushRelEarlyMax}{5\%}
\newcommand{\PushRelEarlyMin}{3\%}
\newcommand{\PushRelLateMax}{49\%}
\newcommand{\PushRelLateMin}{42\%}

\newcommand{\PushRightOtherMax}{89\%}
\newcommand{\PushRightOtherMin}{68\%}

\newcommand{\RandomChangedMax}{34\%}
\newcommand{\RandomChangedMin}{14\%}

\newcommand{\RankNotMax}{8}
\newcommand{\RankNotMin}{4}
\newcommand{\RankRecogMax}{2}
\newcommand{\RankRecogMin}{1}

\newcommand{\ShuffledMax}{8\%}
\newcommand{\ShuffledMin}{6\%}

\newcommand{\SizeN}{224}

\newcommand{\SizeOnsetRight}{36\%}
\newcommand{\SizePreLargeChangedMax}{28\%}
\newcommand{\SizePreLargeMax}{4\%}
\newcommand{\SizePreRandomMax}{4\%}
\newcommand{\SizePreSmallMax}{2\%}

\newcommand{\SportJumpMax}{19}
\newcommand{\SportJumpMin}{7}

\newcommand{\SportLateMax}{90\%}
\newcommand{\SportLateMin}{63\%}

\newcommand{\TextMax}{18\%}
\newcommand{\TextMin}{11\%}

\newcommand{\TightAtJumpMax}{32\%}
\newcommand{\TightAtJumpMin}{16\%}

\newcommand{\TightFoldMax}{25}
\newcommand{\TightFoldMin}{16}

\newcommand{\UnclearMax}{8.7\%}
\newcommand{\UnclearMin}{0.2\%}

\newcommand{\XRelGemma}{90\%}
\newcommand{\XRelLlama}{91\%}
\newcommand{\XRelLlamaSmall}{76\%}
\newcommand{\XRelQwen}{85\%}
\newcommand{\XWordGemma}{92\%}
\newcommand{\XWordLlama}{94\%}
\newcommand{\XWordLlamaSmall}{85\%}
\newcommand{\XWordQwen}{90\%}

%% file: sections/abstract.tex
When a language model answers a factual question, where does its internal state go? We asked four open models (Llama-3.1-8B, Qwen2.5-7B, Gemma-2-9B and Llama-3.2-1B) \NFacts{} factual questions whose answers recur, such as the country a person is a citizen of. At every layer, we recorded the state from which the answer is produced. Questions that receive the same answer converge on a shared region of state space, which we summarise by its centre, the \emph{answer centroid}. The answer becomes readable within a single layer. In every model, the share of questions whose state lies nearest its own answer's centroid jumps at one layer, between \OnsetMin{} and \OnsetMax{} of the way through the network. The centroids follow the answer the model gives rather than the question: the centroid of a country learned from questions about people's citizenship also identifies that country in questions about towns and films. The model also uses the centroids. After the jump, moving the state of a wrongly answered question to the right answer's centroid makes the model give the right answer in \PushGoldMin{}--\PushGoldMax{} of cases late in the network, against \PushRandomMin{}--\PushRandomMax{} for a random move of the same size. Before the jump, moves of the same size almost never work. Near misses, in which the model answers wrongly but picks the right option in multiple choice, can be seen from the onset layer, where the right answer's centroid is already among the nearest to the state. This signal is about as informative as the right answer's rank in the model's final output, and more informative than the model's confidence in its reply.

%% file: sections/intro.tex
\section{Introduction}
\label{sec:intro}

Ask a language model ``What country is An\'{i}bal Ponce a citizen of?'' and, a few dozen layers later, it answers ``Argentina''. On the way, the model's internal state at the last position of the prompt, from which the answer is produced, changes at every layer.

Earlier work reads such intermediate states through the output vocabulary~\citep{nostalgebraist2020interpreting,belrose2023eliciting}, or locates where in the network a fact is retrieved~\citep{geva2023dissecting,meng2022locating}. We ask a geometric question instead. What shape does the change of state take, and do questions with the same answer end up in the same place?

\begin{figure*}[t]
\centering
\includegraphics[width=\textwidth]{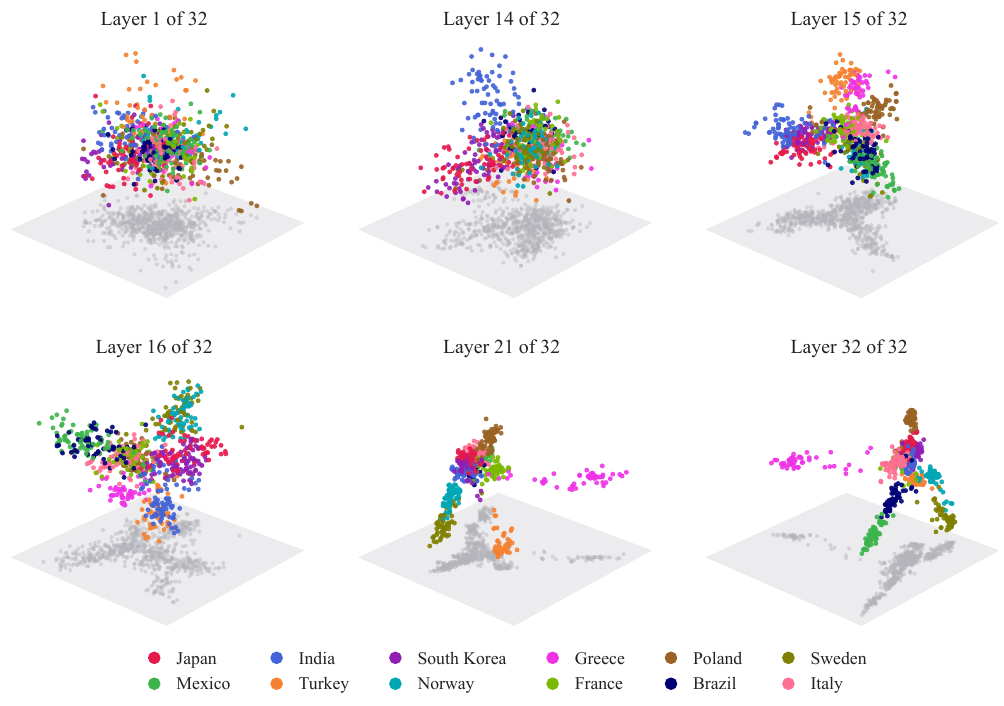}
\caption{\textbf{States of questions at six layers of Llama-3.1-8B.} Each point is one test question about a person, town or film whose answer is one of the twelve countries the model gave most often; colour is the country the model answered. Each panel shows the questions' states at one layer, projected onto the three directions that best separate the twelve countries among the training questions at that layer (found by linear discriminant analysis, LDA). At layer 1 all questions share one cloud, and at layer 14 a few countries, such as India, stand apart. At layer 15, where readability jumps (Section~\ref{sec:step}), most countries separate; by the last layer most form a strand of their own. Each panel is scaled separately; grey dots are shadows on the floor.}
\label{fig:hero}
\end{figure*}

In the four models we study, they do. As a question passes through the network, its state moves into a region that it shares with every other question to which the model gives the same answer. Figure~\ref{fig:hero} shows this in Llama-3.1-8B. Early in the network, questions whose answers are twelve different countries form one tangled cloud. Within a few layers, most of them separate into one strand per country.

To study this, we built \NFacts{} factual questions from Wikidata whose answers recur: 62 countries, cities, languages and sports, each the answer to many questions. We asked them of four open models and recorded the state at every layer. For each answer and layer, we computed the \emph{answer centroid}: the average state of all questions to which the model gave that answer. We then measured, layer by layer, how often a question's state lies nearest the centroid of its own answer. We also tested what the centroids encode, and moved states between centroids to see whether the model uses them. We find the following.
\begin{itemize}\itemsep2pt
\item \emph{The answer becomes readable in a single layer} (Section~\ref{sec:step}). In each model, the share of questions whose state lies nearest its own answer's centroid jumps within one layer. We call it the onset layer. Its depth differs between models.
\item \emph{Centroids follow the model's answer, not the question} (Section~\ref{sec:shared}). A country's centroid learned from questions about people also identifies that country in questions about towns and films. It works just as well when the question is reworded, for example as a casual message. It also survives removing the directions the model uses to write the answer's name, so it is more than the model preparing to output that word. When two models give different answers, each model's state follows its own answer.
\item \emph{The model uses the centroids} (Section~\ref{sec:steer}). After the jump, moving a state to an answer's centroid makes the model give that answer; random moves of the same size rarely do. Before the jump, such moves almost never give the right answer, even when they are as large as moves that work after it.
\item \emph{Near misses are visible from the onset} (Section~\ref{sec:errors}). Some wrong answers are near misses: shown the options, the model picks the right one. In a near miss, the right answer's centroid is already among the nearest to the state at the onset layer. This tells near misses apart from real gaps in knowledge about as well as the model's final output does, and better than its confidence in its reply.
\end{itemize}

%% file: sections/setup.tex
\section{Setup}
\label{sec:setup}

Appendix~\ref{app:details} gives the details needed to reproduce the results.

\subsection{Questions}
\label{sec:questions}
To see whether questions with the same answer reach the same state, we need many different questions for each answer. We built such a set from Wikidata~\citep{vrandecic2014wikidata}: \NFacts{} facts with 62 possible answers of four kinds, namely 24 countries, 16 cities, 12 languages and 10 sports. The facts come in eight types, which Wikidata calls \emph{relations} (Table~\ref{tab:relations}). Several relations share a kind of answer. A country, for example, can answer a question about a person's citizenship, about where a town is, or about where a film is from. The subjects range from famous to obscure, so the models know some facts and not others.

\begin{table}[t]
\centering
\footnotesize
\setlength{\tabcolsep}{3pt}
\begin{tabular}{@{}lll@{}}
\toprule
Answer & Relation & Plain question \\
\midrule
Country & Citizenship & What country is $X$ a citizen of? \\
Country & Town's country & In which country is $X$? \\
Country & Film's country & Which country is the film $X$ from? \\
City & Birthplace & In which city was $X$ born? \\
City & Headquarters & In which city is $X$ headquartered? \\
Language & Language spoken & Which language did $X$ speak or write? \\
Language & Film's language & What is the original language of the film $X$? \\
Sport & Sport played & Which sport does $X$ play? \\
\bottomrule
\end{tabular}
\caption{\textbf{The eight relations.} $X$ is the subject: a person, town, film, company or athlete.}
\label{tab:relations}
\end{table}

Each fact is asked as an open question in up to three wordings: the plain question of Table~\ref{tab:relations} and two rewordings. A language model wrote the rewordings in different styles and was never told the answer. For example, a casual rewording of the plain question ``What country is An\'{i}bal Ponce a citizen of?'' is ``Hi, I'd appreciate knowing which country An\'{i}bal Ponce is a citizen of, if possible?'' The plain question is also asked as multiple choice with four options; this format is used only in Section~\ref{sec:errors}.

\subsection{Models and states}
\label{sec:models}
We study four instruction-tuned models: Llama-3.1-8B-Instruct~\citep{grattafiori2024the}, Qwen2.5-7B-Instruct~\citep{qwen2024qwen25}, Gemma-2-9B-it~\citep{gemma2024gemma} and Llama-3.2-1B-Instruct. Each answers greedily. At the last position of the prompt, where the answer is produced, we record the residual stream after every layer. We call this vector the question's \emph{state} at that layer. \emph{Depth} is the fraction of the model's layers that the state has passed through, from 0 at the input to 1 at the output.

A reply is right or wrong if it names exactly one answer; replies that hedge, refuse or name two answers are set aside (\UnclearMin{}--\UnclearMax{} of replies, depending on the model).

\subsection{Answer centroids}
\label{sec:places}
An answer's \emph{centroid} at a layer is the average state, at that layer, of the questions to which the model gave that answer:
\begin{equation}
c_a = \frac{1}{|Q_a|} \sum_{q \in Q_a} h_q,
\label{eq:place}
\end{equation}
where
\begin{itemize}\itemsep0pt
\item $a$ is an answer, for example Argentina;
\item $Q_a$ is the set of training questions (defined below) that the model answered with $a$, and $|Q_a|$ is their number;
\item $h_q$ is the state of question $q$ at the layer;
\item $c_a$ is the centroid of $a$ at that layer.
\end{itemize}
For example, the centroid of Argentina at layer 15 of Llama-3.1-8B is the average layer-15 state of all training questions that Llama answered with ``Argentina''. These questions may be about a person, a town or a film, in any wording. Centroids are built from the answers the model gives, whether right or wrong. A question that the model wrongly answers with ``Chile'' counts toward the centroid of Chile, so wrong answers can be studied in the same way as right ones.

Centroids are built from questions about one half of the subjects, the \emph{training questions}. Every number we report is measured on questions about the other half, the \emph{test questions}. We split by subject, not by question, because the several questions about one subject are near copies of each other.

\subsection{Readability}
\label{sec:read}
Readability measures how often the answer a model gave can be identified from its state alone. For each test question, we find the centroid nearest to its state, among answers of the same kind:
\begin{equation}
\hat{a}_q = \operatorname*{arg\,min}_{a \in A_q} \big\lVert W(h_q) - W(c_a) \big\rVert,
\label{eq:read}
\end{equation}
where
\begin{itemize}\itemsep0pt
\item $h_q$ is the state of test question $q$ at the layer;
\item $A_q$ is the set of answers of the same kind as the answer the model gave to $q$, for example the 24 countries for a country question;
\item $c_a$ is the centroid of answer $a$ at the same layer;
\item $W$ is a whitening, described below;
\item $\lVert \cdot \rVert$ is the ordinary straight-line distance;
\item $\hat{a}_q$ is the answer whose centroid is nearest.
\end{itemize}
\emph{Readability} at a layer is the share of test questions for which $\hat{a}_q$ is the answer the model actually gave. If states carried no information about answers, readability would be about one over the number of candidate answers: 4\% for countries, 10\% for sports.

Distances are measured after whitening because a few dimensions of the residual stream take much larger values than the rest and would otherwise decide every distance. The whitening $W$ keeps the 128 principal components of the training states at that layer (found by principal component analysis, PCA) and rescales each to unit variance, so that every direction counts equally.

%% file: sections/results.tex
\section{Results}
\label{sec:results}

\subsection{The answer becomes readable in a single layer}
\label{sec:step}

\begin{figure*}[t]
\centering
\includegraphics[width=\textwidth]{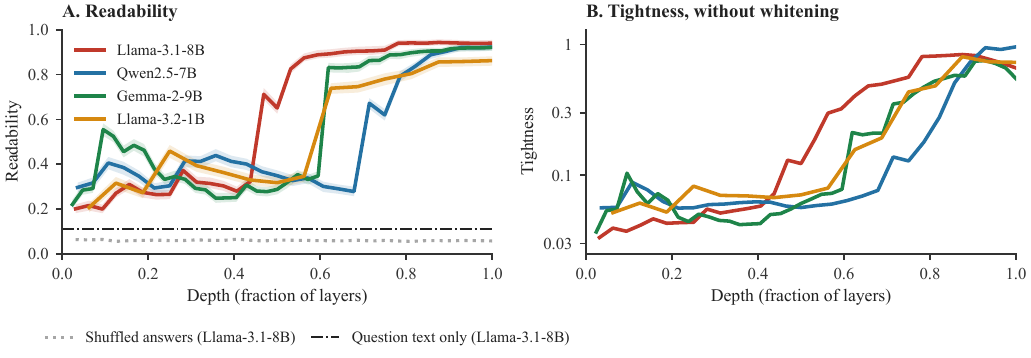}
\caption{\textbf{Readability and tightness by depth.} (A) Readability for the four models. The two horizontal lines are the controls, shown for Llama-3.1-8B; the other models' controls are similar. (B) Tightness for the four models, on a log scale; higher means questions sit closer to their own answer's centroid. Bands are 95\% intervals.}
\label{fig:onset}
\end{figure*}

Readability does not rise smoothly with depth. In every model the curve has a low plateau, then a jump within a single layer, then a high plateau (Figure~\ref{fig:onset}A). In Llama-3.1-8B, readability rises from \JumpFromLlama{} at layer \JumpLayerBeforeLlama{} to \JumpToLlama{} at layer \OnsetLayerLlama{}. The other three models show the same one-layer jump (Table~\ref{tab:models}). After the jump, readability keeps rising to a peak of \PeakLlamaSmall{}--\PeakLlama{}.

We call the layer of the jump the \emph{onset layer}: the first layer at which readability covers half the distance from chance to its peak. To choose it without test questions, we measure this readability on training questions, building centroids from one half of the training subjects and testing on the other (Appendix~\ref{app:details}). The depth of the onset layer, the \emph{onset depth}, differs between models, from \OnsetMin{} to \OnsetMax{} of the way through the network.

Two controls show how much readability our procedure reaches when the state carries no information about the answer.
\begin{itemize}\itemsep2pt
\item \emph{Shuffled answers} keeps every state but credits each question with the answer of a different, randomly chosen fact of the same kind. This measures pure chance, which is low, \ShuffledMin{}--\ShuffledMax{} across models, because each kind has 10 to 24 candidate answers.
\item \emph{Question text only} replaces the model's state with a vector built from the question's spelling alone, without the model.\footnote{The vector counts every sequence of two to four characters in the question, weighted so that sequences common to many questions count less (term frequency--inverse document frequency, TF-IDF), and is compressed to 256 numbers.} This measures how far the wording alone gives the answer away, for example a name that sounds Japanese, and reaches \TextMin{}--\TextMax{}.
\end{itemize}
Readability above both lines is information that the model added to the state.

Before the jump, readability is already \EarlyMin{}--\EarlyMax{}, above both controls; it does not grow toward the jump. This early signal may come from what the model already knows about the subject's name before it retrieves the fact; the question-text control covers only the spelling.

The jump is much smaller for sports, the one kind of answer that is not a name: \SportJumpMin{}--\SportJumpMax{} percentage points, against \NamesJumpMin{}--\NamesJumpMax{} for countries, cities and languages (Section~\ref{sec:limits}).

Readability depends on the whitening. To check the jump in the model's own, unwhitened states, we also measure \emph{tightness}: how closely questions gather around their own answer's centroid. The states grow in size with depth, so we measure this scatter relative to the distance between centroids. At each layer,
\begin{equation}
T = \frac{D_{\text{between}}}{D_{\text{within}}},
\label{eq:tight}
\end{equation}
with
\begin{align*}
D_{\text{between}} &= \frac{1}{|A|} \sum_{a \in A} \lVert c_a - \bar{c} \rVert^2, \\
D_{\text{within}} &= \frac{1}{N} \sum_{q} \lVert h_q - c_{g_q} \rVert^2,
\end{align*}
where
\begin{itemize}\itemsep0pt
\item $A$ is the set of answers of one kind, $|A|$ is their number, and $\bar{c}$ is the average of their centroids;
\item $D_{\text{between}}$ is therefore the average squared distance of a centroid from the centre of all centroids: how far apart the centroids are;
\item the sum in $D_{\text{within}}$ runs over the $N$ test questions of that kind, and $g_q$ is the answer the model gave to question $q$;
\item $D_{\text{within}}$ is therefore the average squared distance of a question from its own answer's centroid: how widely questions scatter;
\item all distances are measured on the unwhitened states.
\end{itemize}
Tightness rises as questions move closer to their own answer's centroid, relative to how far apart the centroids are. We combine the four kinds of answer by weighting each by its number of test questions.

Tightness shows the same event (Figure~\ref{fig:onset}B). From its lowest to its highest value it rises \TightFoldMin{}- to \TightFoldMax{}-fold, and the rise is concentrated at the onset. The two layer steps at the onset, into the onset layer and out of it, carry \TightAtJumpMin{}--\TightAtJumpMax{} of the whole rise on a log scale. If the rise were spread evenly over a model's 16 to 42 layers, two steps would carry only 5\%--13\%. Tightness then keeps rising more slowly for about ten layers. Trained classifiers show no such jump~\citep{he2022a}.

\begin{table}[t]
\centering
\footnotesize
\setlength{\tabcolsep}{2.6pt}
\begin{tabular}{@{}lcccc@{}}
\toprule
 & Llama 8B & Qwen 7B & Gemma 9B & Llama 1B \\
\midrule
Layers & \LayersLlama{} & \LayersQwen{} & \LayersGemma{} & \LayersLlamaSmall{} \\
Accuracy & \AccLlama{} & \AccQwen{} & \AccGemma{} & \AccLlamaSmall{} \\
Onset layer & \OnsetLayerLlama{} & \OnsetLayerQwen{} & \OnsetLayerGemma{} & \OnsetLayerLlamaSmall{} \\
Onset depth & \OnsetLlama{} & \OnsetQwen{} & \OnsetGemma{} & \OnsetLlamaSmall{} \\
\midrule
\multicolumn{5}{@{}l}{\emph{Readability}} \\
Layer before onset & \JumpFromLlama{} & \JumpFromQwen{} & \JumpFromGemma{} & \JumpFromLlamaSmall{} \\
Onset layer & \JumpToLlama{} & \JumpToQwen{} & \JumpToGemma{} & \JumpToLlamaSmall{} \\
Peak & \PeakLlama{} & \PeakQwen{} & \PeakGemma{} & \PeakLlamaSmall{} \\
Standard, 90\% depth & \GivenLateLlama{} & \GivenLateQwen{} & \GivenLateGemma{} & \GivenLateLlamaSmall{} \\
Other relations & \XRelLlama{} & \XRelQwen{} & \XRelGemma{} & \XRelLlamaSmall{} \\
Other wordings & \XWordLlama{} & \XWordQwen{} & \XWordGemma{} & \XWordLlamaSmall{} \\
Answer words removed & \NoWordsLlama{} & \NoWordsQwen{} & \NoWordsGemma{} & \NoWordsLlamaSmall{} \\
True-answer centroids & \GoldLateLlama{} & \GoldLateQwen{} & \GoldLateGemma{} & \GoldLateLlamaSmall{} \\
\midrule
\multicolumn{5}{@{}l}{\emph{Telling near misses from gaps (AUC)}} \\
Centroid rank & \AucRankLlama{} & \AucRankQwen{} & \AucRankGemma{} & \AucRankLlamaSmall{} \\
Output rank & \AucOutRankLlama{} & \AucOutRankQwen{} & \AucOutRankGemma{} & \AucOutRankLlamaSmall{} \\
Confidence & \AucConfMeanLlama{} & \AucConfMeanQwen{} & \AucConfMeanGemma{} & \AucConfMeanLlamaSmall{} \\
\bottomrule
\end{tabular}
\caption{\textbf{Per-model summary.} Accuracy: share of plain test questions answered correctly. Readability rows from ``Standard'' down are at 90\% of depth; the four below ``Standard'' are the tests of Section~\ref{sec:shared}. AUC: Section~\ref{sec:errors}.}
\label{tab:models}
\end{table}

\subsection{Centroids follow the model's answer, not the question}
\label{sec:shared}

\begin{figure}[t]
\centering
\includegraphics[width=\columnwidth]{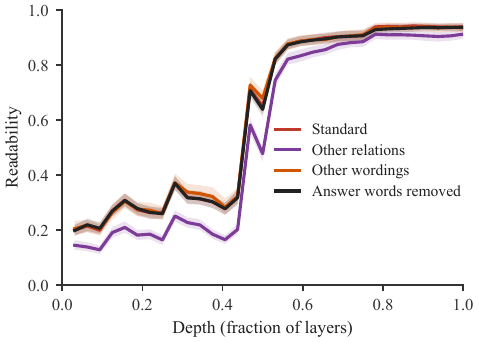}
\caption{\textbf{Readability of Llama-3.1-8B under three changes} (Section~\ref{sec:shared}). The standard, other-wordings and answer-words-removed curves nearly coincide. Bands are 95\% intervals.}
\label{fig:transfer}
\end{figure}

A centroid is built from questions that share an answer. Those questions also share a type of fact, often a wording, and always an output word. We check that the centroid encodes none of these, by changing how the centroids are built or what the state contains and measuring readability again (Figure~\ref{fig:transfer}, Table~\ref{tab:models}). Below, ``standard'' readability is that of Section~\ref{sec:read}; for Llama-3.1-8B it is \GivenLateLlama{} at 90\% of depth.

\paragraph{Other relations.} We built centroids from the training questions of one relation only, measured readability on test questions of the other relations with the same kind of answer, and averaged over the choice of relation. (Sports have a single relation and are left out.) In Llama-3.1-8B, readability at 90\% of depth is \XRelLlama{}, against \NamesLateLlama{} with standard centroids on the same kinds of answer. The centroid of a country is nearly the same whether the question is about a person, a town or a film.

\paragraph{Other wordings.} We built centroids from plain questions only and measured readability on the rewordings, such as the casual example in Section~\ref{sec:questions}. It is \XWordLlama{}, the same as standard.

\paragraph{Answer words removed.} Every question answered with Argentina ends in the same output word, ``Argentina''. We checked whether the centroid is only that word's output direction. We removed from every state its component along the unembedding vectors of all tokens of all 62 answer names. Then we rebuilt the centroids and measured readability again. It stays at \NoWordsLlama{} in Llama-3.1-8B, and removing the same number of random directions does not change it either. The centroid is therefore more than the output direction of its word. Whether it is more than the model's plan to say that word is a different matter. A plan formed in the middle of the network need not lie along the output directions, and our tests cannot tell the two apart.

\paragraph{Different models.} When two models answer the same question differently, the state follows each model's own answer. At the onset layer, each model's state lies nearer its own centroid for its own answer than its own centroid for the other model's answer. This holds for \CrossModelMin{}--\CrossModelMax{} of such questions, across all \CrossModelPairs{} ordered pairs of models. As expected, centroids built from the true answers work less well: at 90\% of depth they identify the true answer in \GoldLateMin{}--\GoldLateMax{} of test questions, against \GivenLateMin{}--\GivenLateMax{} for standard centroids and the given answer.

\subsection{The model uses the centroids}
\label{sec:steer}

Readability shows that the answer can be read from the state. Whether the model itself reads it is a separate question, which we test by moving states. If the model reads the centroids, moving a state to another answer's centroid should change the answer. We took the test questions that the model answered wrongly and whose wrong answer has a centroid (\PushNMin{}--\PushNMax{} per model). For each, we changed its state at one layer to
\begin{equation}
h' = h + \left(c_{\text{right}} - c_{\text{given}}\right),
\label{eq:move}
\end{equation}
where
\begin{itemize}\itemsep0pt
\item $h$ is the question's state at that layer;
\item $c_{\text{given}}$ is the centroid of the wrong answer the model gave;
\item $c_{\text{right}}$ is the centroid of the right answer;
\item $h'$ is the moved state, which replaces $h$ while the model reads the prompt; the model then writes its reply as usual.
\end{itemize}
The centroids are the training centroids of that layer, and the move is made in the model's own, unwhitened states. We made this move at one layer at a time, at seven depths from 20\% to 90\% of the network. As controls, we moved the state by the same distance in a random direction, and toward the centroid of a third, randomly chosen answer of the same kind. We also moved right answers toward another answer's centroid.

\begin{figure*}[t]
\centering
\includegraphics[width=\textwidth]{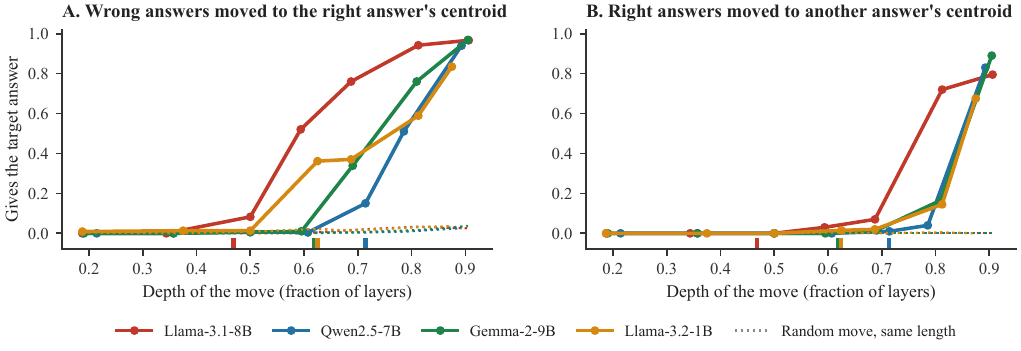}
\caption{\textbf{Moving states to answer centroids.} (A) Wrong answers moved to the right answer's centroid. (B) Right answers moved to another answer's centroid. Solid lines: moves to the centroid; dotted lines: random moves of the same length. The y-axis is the share of questions for which the model then gives the target answer. Short bars under each axis mark each model's onset depth.}
\label{fig:steer}
\end{figure*}

Moves work only after the onset (Figure~\ref{fig:steer}). At every depth before the onset, they correct \PushAllPreMin{}--\PushAllPreMax{} of wrong answers. After the onset, the later the move, the more often it works. At 90\% of depth, it corrects \PushGoldLlama{} of wrong answers in Llama-3.1-8B, \PushGoldQwen{} in Qwen2.5-7B, \PushGoldGemma{} in Gemma-2-9B and \PushGoldLlamaSmall{} in Llama-3.2-1B. Random moves of the same length change the reply in \RandomChangedMin{}--\RandomChangedMax{} of cases, but give the right answer in only \PushRandomMin{}--\PushRandomMax{}. A random move of this size disturbs the reply without directing it.

The effect is not special to the right answer: at 90\% of depth, moves toward a third answer's centroid produce that answer in \PushOtherMin{}--\PushOtherMax{} of cases, and right answers moved toward another answer's centroid switch to it in \PushRightOtherMin{}--\PushRightOtherMax{}.

\paragraph{Why only after the onset?} Centroids move apart with depth, so the moves grow from \PushRelEarlyMin{}--\PushRelEarlyMax{} of the state's length at 20\% of depth to \PushRelLateMin{}--\PushRelLateMax{} at 90\%. Size does not explain why early moves fail. In the main runs, a move of the same size works after the onset and not before. At the last move depth before the onset, the full move is \MatchPreSizeMin{}--\MatchPreSizeMax{} of the state's length and corrects \MatchPreRightMin{}--\MatchPreRightMax{} of wrong answers. At the first move depth at or after the onset, a half move of similar size (\MatchPostSizeMin{}--\MatchPostSizeMax{}) corrects \MatchPostRightMin{}--\MatchPostRightMax{}. This holds in all four models.

We also made early moves much larger in Llama-3.2-1B (onset layer 10). At layers 3, 6 and 8, we scaled the move from the given to the right centroid up to 17\% and 44\% of the state's length. At the onset layer, a 17\% move corrects \SizeOnsetRight{} of the \SizeN{} wrong answers. Before the onset, a 17\% move corrects at most \SizePreSmallMax{}, and a 44\% move at most \SizePreLargeMax{}, the same as a random 44\% move (\SizePreRandomMax{}). The model does respond to these large early moves. They change the reply in up to \SizePreLargeChangedMax{} of cases, almost never to the right answer.

Before the onset, questions scatter far more widely than the centroids are apart (low tightness, Figure~\ref{fig:onset}B). What separates them is partly readable: \JumpFromMin{}--\JumpFromMax{} at the layer before the onset, against \ShuffledMin{}--\ShuffledMax{} with shuffled answers. Moving along it works like a random move. A possible reason is that the answer reaches the last position only at the onset~\citep{geva2023dissecting}, so anything added before then is overwritten when it arrives. We have not tested this.

Late moves are large and close to the output, and could write the answer into the state directly. The stronger evidence that the model reads the centroids is the onset itself and the failure of random moves of the same length.

\subsection{Near misses are visible from the onset}
\label{sec:errors}

Some wrong answers are \emph{near misses}: the model gets the open question wrong but picks the right option when the same question is asked as multiple choice. We count a wrong answer as a near miss only if the model picks the right option in at least three of the four option orders. This holds for \ErrPosLlama{} of wrong answers in Llama-3.1-8B, \ErrPosQwen{} in Qwen2.5-7B, \ErrPosGemma{} in Gemma-2-9B and only \ErrPosLlamaSmall{} in Llama-3.2-1B. We call the other wrong answers \emph{gaps}. Does the state at the onset already tell the two apart?

For each wrong answer to a plain test question, we compare three scores. The \emph{centroid rank} is the rank of the right answer's centroid when the centroids of the question's kind are ordered by distance from the state, 1 being the nearest. We take it at the onset layer of Section~\ref{sec:step}. The \emph{output rank} is the same ranking made from the model's output probabilities for the first token of each candidate answer. The model's \emph{confidence} is its average log-probability per token of its own reply; it is the only score that does not need the right answer. We measure how well each score separates near misses from gaps by the area under the receiver operating characteristic curve (AUC). This is the probability that a randomly chosen near miss scores better than a randomly chosen gap; 0.5 is chance and 1 is perfect. No score has parameters fitted to these answers.

\begin{figure}[t]
\centering
\includegraphics[width=\columnwidth]{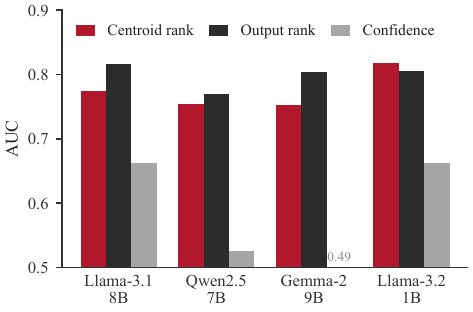}
\caption{\textbf{Telling near misses from gaps.} AUC of each number among wrong answers: the centroid rank at the onset layer, the output rank, and the model's confidence in its reply. The axis starts at chance (0.5), so Gemma-2-9B's confidence (0.49) has no bar.}
\label{fig:errors}
\end{figure}

The centroid rank separates near misses from gaps about as well as the output rank: AUC \AucRankMin{}--\AucRankMax{}, against \AucOutRankMin{}--\AucOutRankMax{} (Figure~\ref{fig:errors}). Both do better than confidence, \AucConfMeanMin{}--\AucConfMeanMax{}. For near misses, the right answer's centroid has a median rank of \RankRecogMin{}--\RankRecogMax{}, depending on the model; for gaps, \RankNotMin{}--\RankNotMax{}.

In a near miss, the right answer is already a close runner-up at the onset, often just behind the answer the model gave. The final output keeps this information; the model's confidence in its reply does not.

The result is not an effect of the kind of answer. Kinds differ in their share of near misses and their number of candidates, so we repeated the test within countries, the largest kind. There, the centroid rank still beats confidence in every model: AUC \CountryAucRankMin{}--\CountryAucRankMax{}, against \CountryAucConfMin{}--\CountryAucConfMax{}.

%% file: sections/related.tex
\section{Related work}
\label{sec:related}

\paragraph{Reading intermediate layers.}
The logit lens~\citep{nostalgebraist2020interpreting} decodes an intermediate state through the model's output layer, and the tuned lens~\citep{belrose2023eliciting} learns a decoder for each layer. Both show predictions sharpening with depth. Future Lens~\citep{pal2023future} and Patchscopes~\citep{ghandeharioun2024patchscopes} read richer content from a single hidden state. All of these read the state through the output vocabulary. Our centroids are fitted from the states themselves, and reading them does not need the answer words' output directions.

\paragraph{How depth is used.}
Early layers gather and detokenise, middle layers build abstract features, and late layers refine a prediction~\citep{lad2024the,csordas2025do,gupta2025how,cheng2024emergence,skean2025layer}. Correct and incorrect predictions follow similar confidence curves across layers~\citep{kim2025on}. In factual recall, information about the relation and then the subject moves to the last position, where attention heads extract the attribute~\citep{geva2023dissecting}. Feed-forward layers act as key--value memories~\citep{geva2020transformer}, and middle-layer feed-forward modules at the subject's last token are causally decisive~\citep{meng2022locating}. Our onset depth marks, for a whole population of questions, the depth at which each model's answer becomes available at the last position.

\paragraph{Representation geometry.}
Concepts in language models have structured geometry~\citep{park2024the,engels2024not,gurnee2023language,gurnee2026when,marks2023the}, and middle layers share content across languages and modalities~\citep{wendler2024do,wu2024the}. In trained classifiers, classes separate by a roughly constant factor per layer~\citep{he2022a}; language models move partway toward neural collapse, in which states gather at their class means, at their top layer~\citep{wu2024linguistic}. The residual stream also contains stable regions in which small changes leave the prediction unchanged~\citep{janiak2024characterizing}, although, almost surely, no two prompts map to the same state~\citep{nikolaou2025language}. Centroids fit both: questions can gather in the directions that carry the answer while staying distinct in others.

\paragraph{What models know but do not say.}
Internal states often encode more than the output shows~\citep{orgad2024llms,gekhman2025insideout,ruscio2026on}. On knowledge benchmarks such as MMLU and ARC, however, this gap is near zero~\citep{park2025bridging}, and internal states mostly reflect what the model recalls rather than what is true~\citep{cheang2026do}. False few-shot demonstrations can override a correct internal answer at a critical layer~\citep{halawi2023overthinking}, and users' suggestions shift answers~\citep{sharma2023towards}. In multiple choice, attention in a few middle layers selects the answer letter, and a sparse set of heads then raises its probability~\citep{wiegreffe2024answer}. The winning option is chosen by its content before it is bound to its letter~\citep{wong2026when}, and models carry a prior over letters~\citep{zheng2023large}.

\paragraph{Steering.}
Adding a direction to the residual stream changes a model's behaviour~\citep{panickssery2023steering,li2023inferencetime}. The moves here are a special case: the direction is the difference between two measured centroids, applied to one question at one layer, and the control is a random move of the same length.

%% file: sections/limits.tex
\section{Limitations}
\label{sec:limits}

The questions are short and factual, and their answers come from a closed set of 62 countries, cities, languages and sports. Three of these four kinds are names. For the one kind that is not, sports, the jump is much smaller (Section~\ref{sec:step}) and readability at 90\% of depth is lower (\SportLateMin{}--\SportLateMax{} against \NamesLateMin{}--\NamesLateMax{}). Sports have at most ten answers, all from one relation, so we cannot tell whether this reflects the kind of word or the smaller set. Centroids for open-ended answers, multi-step reasoning or long contexts may look different.

We studied four instruction-tuned models between 1 and 9 billion parameters, with one prompt format each and greedy answers. Readability and tightness both measure where states sit, not the computation that moves them there. Readability jumps within one layer, but tightness keeps rising for about ten layers after the onset, so the convergence itself is not instantaneous.

Wikidata labels contain errors. We removed one unreliable relation--answer pair after a manual check, and a small number of wrong labels surely remain. The rewordings were written by a language model and filtered automatically.

%% file: sections/conclusion.tex
\section{Conclusion}
\label{sec:conclusion}

We followed the state from which four language models answer short factual questions, layer by layer. For the first half to two thirds of each network, the state carries only a weak trace of the answer, and moving it almost never changes the answer to the right one. Then, at one layer, the answer arrives. From that layer on, every question with the same answer sits near one centroid, whatever was asked, and moving a state to another centroid changes the reply.

Wrong answers are placed the same way, so at the onset a near miss shows up as a right answer that came second. We have not tested larger models, or open-ended answers, which have no fixed set of centroids; these are the obvious next cases.

%% file: sections/ack.tex
This work was inspired by Goodfire's Neural Geometry series~\citep{goodfire2026the}. The questions come from Wikidata~\citep{vrandecic2014wikidata}. Claude, GPT and Fable helped with literature search, code review and drafting. Every number in the paper is computed from the recorded data by the released code.

%% file: sections/appendix.tex
\section{Details}
\label{app:details}

\paragraph{Questions.} Subjects were sampled from three popularity bands by number of Wikipedia language editions (1--4, 5--19, and 20 or more), roughly a third from each, because models know popular facts far better than rare ones~\citep{mallen2022when}. Every subject has one answer, a unique English name and no namesake of the same kind. We dropped Hindi as the language a person spoke after a manual check found its labels unreliable. A rewording was kept only if it named the subject exactly, asked a question, mentioned no candidate answer or its demonym, and differed from the plain question; \NRewrites{} passed. Multiple-choice questions show four options in all four rotations of their order. In total each model answers \NPromptsAnalysed{} prompts, each in its own chat format.

\paragraph{States and grading.} States are recorded before the final normalisation, exactly as the model computes them in bfloat16, at the last prompt position before the first layer and after every layer. A reply is graded by its first line, matched against hand-curated names for each answer (for example ``China'' or ``Mandarin''). A single answer outside the 62 candidates counts as wrong.

\paragraph{Centroids and readability.} A centroid is built only for an answer given to at least five training questions about at least three different subjects. Readability counts only test questions whose given answer has a centroid. The whitening keeps the 128 leading principal components of the training states at each layer and divides each by its standard deviation, after shrinking each variance 10\% toward the average variance so that directions with very little variance are not blown up. When a result is reported at a given depth, we use the nearest layer.

\paragraph{Onset layer.} We split the training subjects in half again, build centroids from one half, and measure readability on the other at every layer. The onset layer is the first layer at which this readability covers half the distance from chance to its highest value over layers, where chance is the average, over questions, of one over the number of candidate answers of the question's kind.

\paragraph{Uncertainty.} Questions about one subject are not independent, so we resample subjects: we draw the test subjects with replacement 300 times, recompute the result each time, and report the middle 95\% of the results.

\paragraph{Controls.} Shuffled answers are permuted between facts of the same kind, all wordings of a fact keeping one answer; readability is averaged over five shuffles. For the question-text control, the TF-IDF vectors of character sequences of length two to four are reduced to 256 dimensions by a truncated singular value decomposition fitted on training questions.

\paragraph{Tightness.} $D_{\text{between}}$ and $D_{\text{within}}$ are computed for each kind of answer separately and each is summed over kinds, weighting each kind by its number of test questions, before taking their ratio.

\paragraph{Centroids follow the answer.} For other relations, each test question's readability is averaged over the centroids built from each other relation of its kind. For answer words removed, each unembedding vector is first multiplied by the final normalisation's scale, the vectors are turned into an orthonormal basis, and each state is projected onto the space at right angles to that basis; the whitening is then refitted. The random control removes a random orthonormal basis of the same size. True-answer centroids are built like ordinary centroids but group training questions by their true answer. The cross-model comparison uses plain test questions at each model's onset layer.

\paragraph{Moves.} The seven depths are 20\%, 35\%, 50\%, 60\%, 70\%, 80\% and 90\%, each rounded to the nearest layer. The move is added once, at the last prompt position, while the model reads the prompt; generation then proceeds without changes. The third answer, and the other answer for right answers, are drawn at random from answers of the same kind that have a centroid. Right answers are a random sample of 200 per model. Wrong answers whose given answer has no centroid are excluded from the numbers in Section~\ref{sec:steer}.

\paragraph{Size check.} The check uses Llama-3.2-1B on a laptop and the same \SizeN{} wrong answers as the main moves (script: \texttt{wisdom/size\_check.py}). Each move is rescaled to a fixed share of the median length of test states at its layer; random moves use a fixed random direction per question and layer.

\paragraph{Near misses.} The centroid rank is computed in whitened states, for wrong answers whose right answer has a centroid; all three numbers are computed on these same wrong answers. The output rank uses the model's log-probability, at the first position of its reply, of the first token of each candidate's names, taking the best name per candidate. As a check, logistic regressions on three geometric features (distance to the right centroid, its rank, and the margin between the two nearest centroids) and on the reply's log-probability and length, scored by five-fold cross-validation with each subject kept in one fold, give AUC \AucGeoMin{}--\AucGeoMax{} against \AucConfMin{}--\AucConfMax{}.